\documentclass[11pt]{article}

\usepackage[final]{acl}

\usepackage{times}
\usepackage{latexsym}

\usepackage[T1]{fontenc}

\usepackage[utf8]{inputenc}

\usepackage{microtype}

\usepackage{inconsolata}

\usepackage{graphicx}

\usepackage{amsmath}
\usepackage{booktabs}
\usepackage{subcaption}
\usepackage{multirow}
\usepackage{algorithm}
\usepackage{algorithmic}
\title{Entropy-Aware Token Rejection for Improving Speculative Decoding}

\author{
  \textbf{Tiancheng Su}\textsuperscript{1}
  \qquad
  \textbf{Meicong Zhang}\textsuperscript{2}
  \qquad
  \textbf{Guoxiu He}\textsuperscript{2}\thanks{Corresponding author.}
  \\
  \textsuperscript{1} Wuhan University, Wuhan, China
  \\
  \textsuperscript{2} East China Normal University, Shanghai, China
  \\
  \texttt{tianchengsu@whu.edu.cn},
  \texttt{mczhang@stu.ecnu.edu.cn},
  \texttt{gxhe@fem.ecnu.edu.cn}
}

\begin{document}
\maketitle
\begin{abstract}
Speculative decoding (SD) accelerates large language model (LLM) inference by using a lightweight draft model to propose tokens and a stronger target model to verify them. However, standard SD is mainly designed for acceleration, and its output quality is typically constrained by the target model. In this work, we propose Entropy-Aware Speculative Decoding (EASD), a lightweight and training-free extension of SD that improves reasoning quality through token-level entropy-guided rejection. EASD detects cases where both draft and target models exhibit high uncertainty while strongly overlapping in their top predictions. In such uncertain-agreement cases, EASD rejects the aligned token and resamples from the target distribution, preventing low-confidence errors from propagating. Experiments on challenging reasoning benchmarks show that EASD consistently improves accuracy over standard SD and reward-guided variants while maintaining comparable inference efficiency. Notably, EASD can surpass the standalone performance of the target model, suggesting that speculative decoding can serve not only as an acceleration method but also as an effective mechanism for improving reasoning quality. The code is available at  \url{https://github.com/ECNU-Text-Computing/EASD}.
\end{abstract}

\section{Introduction}

\begin{figure}[h]
\centerline{\includegraphics[width=0.5\textwidth]{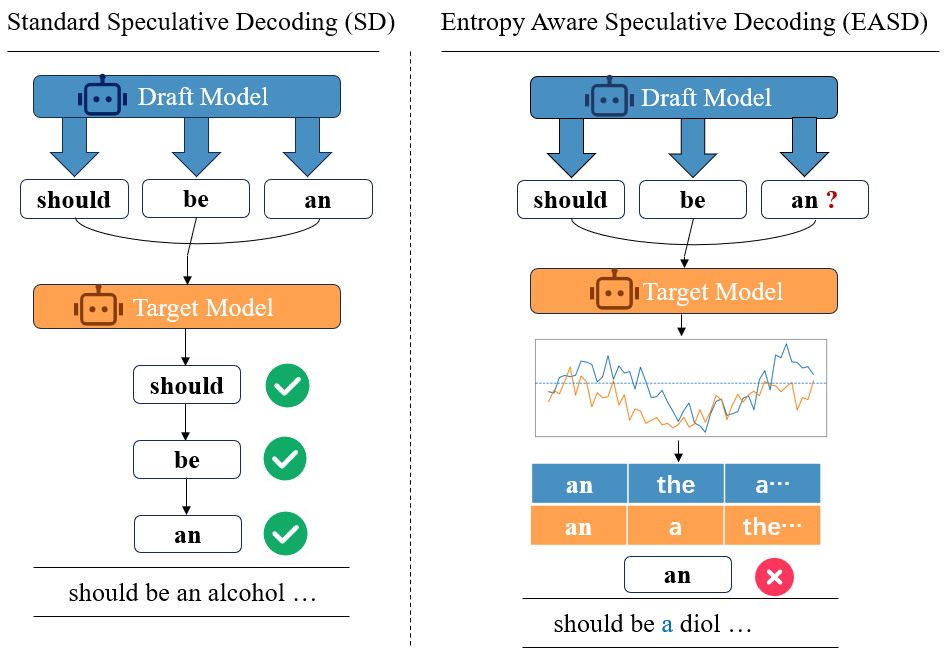}}%
\caption{Standard speculative decoding (SD) verifies tokens against the target model and thus tends to treat verified outputs as reliable. For same-family draft/target pairs, high entropy with strong top-k overlap can indicate "uncertain agreement"; EASD detects this case, rejects the aligned token, and resamples from the target to prevent error propagation: \textit{e.g.}, replacing "an" in "should be an …" with "a", shifting "an alcohol" to the correct "a diol."}
\label{fig:intro}
\end{figure}

Large language models (LLMs), such as ChatGPT \citep{achiam2023gpt}, Qwen \citep{bai2023qwen}, and LLaMA \citep{touvron2023llama}, have achieved strong performance across a wide range of natural language processing tasks. However, their autoregressive decoding process requires a full forward pass for each generated token, leading to high latency and computational cost, especially for long-form reasoning and large-scale deployment \citep{patterson2021carbon,frantar2022gptq,lin2024awq}. Speculative decoding (SD) \citep{chen2023accelerating} addresses this problem by using a smaller draft model to generate candidate tokens, which are then verified in parallel by a stronger target model. This reduces the number of target-model invocations while largely preserving generation quality. Subsequent methods, including Fast Inference \citep{leviathan2023fast}, Medusa \citep{cai2024medusa}, Hydra \citep{ankner2024hydra}, and EAGLE \citep{li2024eagle,li2024eagle2,li2025eagle}, further improve SD through multi-token verification, parallel decoding heads, and adaptive candidate generation.

Despite its efficiency, standard SD remains primarily constrained by the target model, since accepted tokens are ultimately judged according to the target distribution. This makes SD effective for acceleration, but less capable of correcting uncertain target-model decisions. To improve output quality beyond acceleration, recent studies introduce reward models into SD-style decoding \citep{li2025reward,liao2025reward}. These methods can guide token or sequence selection using preference or correctness signals, but they require additional model inference and often depend on task-specific training data \citep{gao2023scaling,lambert2024rewardbench}. This increases computational overhead and limits generalization across tasks, datasets, and model pairs.

In this work, we propose Entropy-Aware Speculative Decoding (EASD), a training-free enhancement to SD. EASD is motivated by the observation that errors often arise when both the draft and target models are uncertain but still agree on a plausible token. Such cases may reflect not confident correctness, but uncertain agreement. This phenomenon is especially important in reasoning tasks, where an early token-level mistake can redirect the entire subsequent solution path. EASD detects this pattern using two signals: the entropy of the draft and target distributions, and the overlap between their top candidate tokens. When both models have high entropy and strong top-$N$ overlap, EASD rejects the aligned token and resamples from the target distribution. This simple intervention reduces the propagation of low-confidence tokens and improves reasoning reliability without introducing additional reward models or training procedures.

We evaluate EASD on challenging reasoning benchmarks, including Olympia \citep{he2024olympiadbench}, MATH500 \citep{hendrycks2021measuring}, and GPQA-Diamond \citep{rein2024gpqa}. The results show that EASD improves generation accuracy over standard SD and reward-guided baselines while preserving comparable inference efficiency. More importantly, EASD often outperforms the standalone target model, demonstrating that speculative decoding can be used not only for acceleration but also for quality improvement. This suggests that draft-target interaction can provide useful uncertainty signals beyond simple verification.

Our main contributions are summarized as follows:

\begin{itemize}
    \item We propose EASD, a lightweight and training-free extension of speculative decoding based on token-level entropy signals.
    \item We introduce an entropy-guided rejection mechanism that targets high-uncertainty agreement between draft and target models to reduce error propagation.
    \item Experiments on multiple reasoning benchmarks show that EASD improves accuracy over existing SD variants and can surpass the standalone target model.
\end{itemize}

\section{Related Work}

\subsection{Collaboration Between Large and Small Language Models}

Collaboration between large and small language models has become an important direction for balancing inference efficiency and generation quality. Existing approaches often use small models as lightweight assistants for query reformulation, intermediate annotation, or task routing before invoking larger models \citep{chen2025survey,shen2024learning,yang2025speculative,pan2025specreason}. While effective, these methods usually connect models at a coarse-grained level and rely on prompt-based information transfer, limiting deeper collaboration during decoding.

Recent work explores finer-grained interaction by interleaving or fusing token distributions from multiple models \citep{hao2025dynamic,bian-etal-2025-ptoco}. Such methods allow models to participate jointly in token-level generation, but efficient and stable integration remains challenging. In contrast, EASD focuses on a simple decoding-level collaboration mechanism: the draft model does not merely accelerate generation, but also provides uncertainty signals that help identify unreliable target-model decisions.

\subsection{Speculative Decoding}

Speculative decoding accelerates LLM inference by allowing a draft model to propose candidate tokens and a target model to verify them \citep{chen2023accelerating}. Many follow-up studies improve candidate generation, verification efficiency, and acceptance strategies \citep{sun2024optimal,ankner2024hydra,cai2024medusa,li2025eagle,li2025reward}. These methods mainly aim to reduce target-model computation while maintaining the target model's output distribution.

Recent studies further suggest that speculative decoding can be extended beyond acceleration. For example, partially aligned or deliberately biased draft models may introduce useful preferences or task-specific signals, enabling the system to improve generation quality under certain conditions \citep{bachmann2025judge,liao2025reward}. However, many such approaches rely on trained reward or evaluation models, which increases cost and reduces portability. EASD instead uses entropy and top-candidate overlap as training-free signals, making it easier to integrate into existing SD pipelines.

\subsection{Entropy-Based Inference}

Entropy is widely used as an uncertainty measure in language generation \citep{wang2025beyond}. Prior studies have used entropy to adjust sampling temperature, guide beam search, set adaptive thresholds, or dynamically route computation across models \citep{simonds2025entropy,yang2025less,zhang2024edt,qiu2024entropy}. These methods show that entropy can help balance efficiency, exploration, and reliability during inference.

Recent studies have also incorporated uncertainty or confidence into speculative decoding. HeteroSpec allocates verification effort according to candidate uncertainty to improve decoding efficiency \citep{liu2026heterospec}. SpecGuard performs step-level verification using attention-based grounding and log-probability-based confidence signals \citep{purohit2026tokens}, while MARS introduces margin-aware speculative verification based on the target model's decision stability \citep{song2026mars}. In contrast, EASD combines entropy with draft--target top-$n$ overlap to identify uncertain agreement and improve reasoning accuracy.

\section{Entropy-Aware Speculative Decoding}
\label{sec:method}

In this section, we present \textbf{Entropy-Aware Speculative Decoding (EASD)}, a novel, training-free extension of standard speculative decoding. EASD enhances the reliability and quality of generated outputs by leveraging entropy-based signals from both the draft and target models. It dynamically adjusts the target model's token selection under high uncertainty and distributional similarity, using a targeted penalty mechanism. This approach mitigates error propagation from the draft model, improving decoding efficiency and output coherence without additional training.

\subsection{Formulation} 
The key idea of speculative decoding is to generate a draft sequence speculatively and then accept as many prefix tokens as possible based on a verification step. This method leverages the fact that draft models can often predict correctly for several steps, especially when they approximate the target model well.

Let \( p_t(\cdot \mid \mathbf{x}) \) denote the conditional probability distribution of the target model given a prefix \( \mathbf{x} \), and \( p_d(\cdot \mid \mathbf{x}) \) denote that of the draft model. The draft model generates a speculative sequence of candidates \( c_1, c_2, \dots, c_k \), where each \( c_i \) is sampled as: \(
c_i \sim p_d(\cdot \mid \mathbf{x}, c_1, \dots, c_{i-1}).
\)
Here, \( i \) is the speculation length, typically chosen based on model sizes and desired speedup.

During the verification phase, the target model computes probabilities for all drafted tokens in parallel. Each candidate \( c_{i} \) is either accepted or rejected based on the agreement between the two models. Formally, token \( c_{i} \) is accepted only if all previous candidates are also accepted and
\begin{equation}
\label{eq:1}
\epsilon_{i} \leq \frac{p_{t}(c_i \mid \mathbf{x}, c_1, \dots, c_{i-1})}{p_d(c_i \mid \mathbf{x}, c_1, \dots, c_{i-1})}, \qquad \epsilon_{i}\sim\mathcal{U}(0,1).
\end{equation}
Specifically, $\epsilon_{i}$ is sampled uniformly from $(0,1)$. 
If the probability of token $c_i$ under the target model exceeds that under the draft model, the token is accepted. 
Otherwise, it is accepted with a probability that decreases as the gap between the two probabilities increases. 
If the token is rejected, the procedure resorts to direct sampling from the target model, drawing a correction token from 
\( p_t(\cdot \mid \mathbf{x}, c_1, \dots, c_{i-1}) \).

\subsection{Overview}
\label{subsec:overview}

Speculative decoding accelerates LLM inference by using a smaller draft model to propose candidate tokens, which the larger target model verifies through a single forward pass. At step $t$, the draft model samples a token from its distribution $P_d^{(t)}$ over $\mathcal{V}$, and the target model accepts or rejects the token based on agreement. This reduces computation but risks potential errors. The risk is amplified when both the draft and target models are uncertain, as they may align in their uncertainty and lead the verification step to accept suboptimal tokens. 

EASD improves this by integrating entropy-based uncertainty and distributional overlap. It monitors entropy in both models and overlap in their top-$n$ tokens. When uncertainty is high and overlap is significant, EASD penalizes the target model's probability for the draft's top token, promoting alternative selections. This enhances robustness without altering model training.

\subsection{Entropy Computation}
\label{subsec:entropy}

To measure uncertainty at step $t$, we use Shannon entropy for the draft model's distribution $P_d^{(t)}$:
\begin{equation}
H_d^{(t)} = - \sum_{i \in \mathcal{V}} P_d^{(t)}(i) \log P_d^{(t)}(i),
\label{eq:draft_entropy}
\end{equation}
and similarly for the target model:
\begin{equation}
H_t^{(t)} = - \sum_{i \in \mathcal{V}} P_t^{(t)}(i) \log P_t^{(t)}(i).
\label{eq:target_entropy}
\end{equation}
High entropy reflects greater uncertainty, corresponding to a flatter distribution, whereas low entropy indicates higher confidence with a more peaked distribution.

\subsection{Triggering Conditions for Dynamic Penalty}
\label{subsec:trigger_conditions}
 
EASD applies a penalty only when two specific conditions are met: high entropy and top-$n$ overlap. This combination identifies scenarios where the draft model is likely to mislead the target model. The rationale is that when models from the same family exhibit high uncertainty yet similar token distributions, it often indicates ambiguous decision points where errors are prone to propagate.

{
\renewcommand{\labelenumi}{(\alph{enumi})}
\begin{enumerate}
    \item \textbf{High-Entropy Condition:} Both models show high uncertainty if:
    \begin{equation}
    H_d^{(t)} > \tau_H \quad \text{and} \quad H_t^{(t)} > \tau_H,
    \label{eq:high_entropy}
    \end{equation}
    where $\tau_H$ is a tunable threshold. High entropy indicates a flat probability distribution, meaning the model lacks a clear preference, which makes its decision susceptible to minor perturbations.

    \item \textbf{Top-$n$ Overlap Condition:} The overlap ratio between top-$n$ token sets $T_d^n$ (draft) and $T_t^n$ (target) is:
    \begin{equation}
    \text{Overlap}(T_d^n, T_t^n) = \frac{|T_d^n \cap T_t^n|}{n},
    \label{eq:overlap_ratio}
    \end{equation}
    (\textit{e.g.}, $n=5$). The condition triggers if
    $\text{Overlap} \geq \tau_O$, where $\tau_O \in [0, 1]$.
    This assesses distributional similarity, highlighting potential alignment issues.
\end{enumerate}
}

\subsection{Application of the Dynamic Penalty}
\label{subsec:penalty_application}

If both conditions are met, EASD rejects the draft's proposed token $t_d$ by setting:
\begin{equation}
P_t^{(t)}(t_d) = 0,
\label{eq:penalty_application}
\end{equation}
and renormalizing:
\begin{equation}
P_t^{(t)}(i) = \frac{P_t^{(t)}(i)}{\sum_{j \neq t_d} P_t^{(t)}(j)}, \quad \forall i \neq t_d.
\label{eq:renormalization}
\end{equation}

This forces the target model to sample alternatives, reducing error risks and improving coherence. 

\begin{figure}[h]
\centerline{\includegraphics[width=0.5\textwidth]{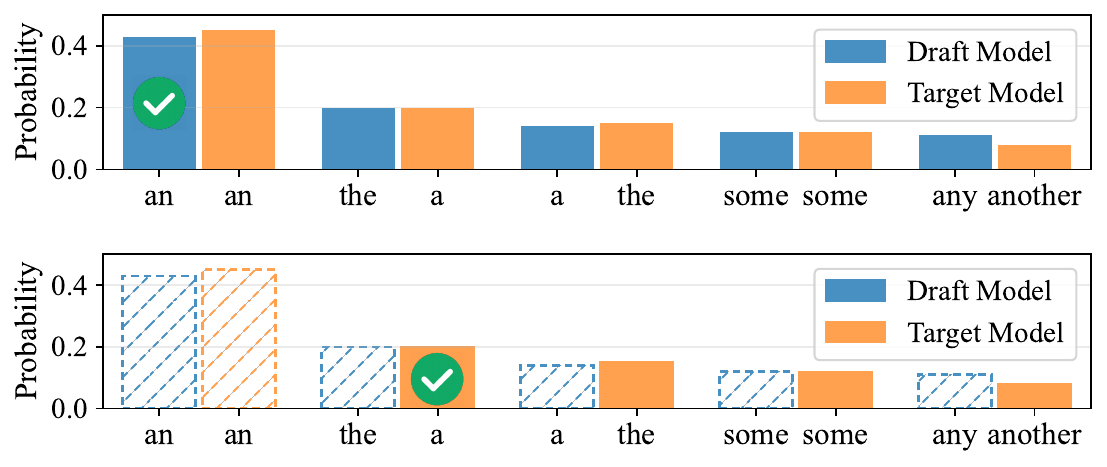}}%
\caption{In the upper panel, standard speculative decoding accepts the draft token "an" at step $i$ because the target model assigns it higher probability. In the lower panel, EASD detects a high-entropy and high-overlap regime, rejects the aligned "an" as a likely error, and resamples from the remaining target distribution.}
\label{fig:case}
\end{figure}

As illustrated in Figure \ref{fig:case}, standard speculative decoding accepts a drafted token whenever it is sufficiently supported by the target model under the acceptance test in Eq.~\ref{eq:1}. In the upper panel, the drafted token "an" receives a higher probability under the target model than under the draft model, so it is accepted and the decoding continues along that branch. In the lower panel, EASD detects that both models are in a high-entropy regime (Eq.~\ref{eq:high_entropy}) while their top-$n$ predictions exhibit high overlap (Eq.~\ref{eq:overlap_ratio}). Under these conditions, EASD applies the dynamic penalty in Eq.~\ref{eq:penalty_application} by setting $P_t^{(t)}(t_d)=0$ and renormalizing Eq.~\ref{eq:renormalization}, which forces sampling from the remaining target distribution instead of committing to the aligned draft token.

\subsection{Threshold Selection}
\label{subsec:threshold_selection}

Proper selection of the thresholds $\tau_H$ (entropy) and $\tau_O$ (top-$n$ overlap) is critical for the performance of EASD, as they directly determine when the dynamic penalty is applied. We adopt a systematic, data-driven procedure to set these thresholds without introducing additional training. 

\textbf{Entropy Threshold $\tau_H$.} 
For each validation sample $s \in \mathcal{D}_{\mathrm{val}}$ and decoding step $t \in \{1,\dots,T_s\}$, the target model produces a probability distribution $P_{s,t}(\cdot)$ over the vocabulary $\mathcal{V}$. The Shannon entropy is defined as
\begin{equation}
H_{s,t} = - \sum_{v \in \mathcal{V}} P_{s,t}(v) \, \log P_{s,t}(v).
\end{equation}
We collect all entropy values across the validation set into
\begin{equation}
\mathcal{E} = \{ H_{s,t} : s \in \mathcal{D}_{\mathrm{val}}, \; t = 1,\dots,T_s \}.
\end{equation}
Let $p \in (0,1)$ denote the proportion of high-entropy steps considered ($p=0.05$). 
We identify the subset $\mathcal{S}_p \subseteq \mathcal{E}$ corresponding to the top $p$ fraction of entropy values, \textit{i.e.}, 
$|\mathcal{S}_p| = \lceil p \cdot |\mathcal{E}| \rceil$. 
The entropy threshold is then defined as the mean entropy of these uncertain steps:
\begin{equation}
\tau_H = \frac{1}{|\mathcal{S}_p|} \sum_{H_{s,t} \in \mathcal{S}_p} H_{s,t}.
\end{equation}

To ensure comparability with recent methods under a consistent hyperparameter selection protocol, the resulting value is further mapped to its nearest candidate in $\{0.5, 1.0, 1.5, 2.0\}$ and used as the final entropy threshold $\tau_H$.

This construction ensures that $\tau_H$ reflects a representative level of uncertainty by focusing on the most ambiguous decoding steps, while maintaining sufficient statistical robustness.

\textbf{Top-$n$ Overlap Threshold $\tau_O$.}
Complementary to the entropy-based criterion, the overlap threshold $\tau_O$ regulates cases where the draft and target models produce highly similar candidate distributions. We fix $\tau_O = 0.8$, meaning that when at least four out of the top-$5$ tokens coincide between the two models, the dynamic penalty is triggered. This setting reflects the intuition that excessive alignment between the draft and target models reduces the diversity and corrective potential of speculative decoding, and thus requires additional penalization to prevent redundant acceptance.

\section{Experiments}

\subsection{Experimental Settings}

\textbf{LLMs.} To assess the performance of EASD, we employ Qwen2.5-7B-Instruct, Qwen2.5-32B-Instruct, and Qwen2.5-72B-Instruct models as our draft and target models. For experiments involving PRM, we select the Skywork-o1-Open-PRM-Qwen-2.5-1.5B model.

\textbf{Datasets.} We evaluate our method on a diverse set of reasoning tasks, including OlympiadBench \citep{he2024olympiadbench}, MATH500 \citep{hendrycks2021measuring}, AIME24 , AMC23 , GPQA-Diamond \citep{rein2024gpqa}, and Minerva Math \citep{lewkowycz2022solving}.

\textbf{Baselines.} We categorize the baselines into five groups:  
\textbf{(1) Target model only.} This baseline executes the target model independently, which typically results in higher computational cost compared to EASD.  
\textbf{(2) Draft model.} This category consists of common test-time scaling techniques built upon the draft model, including draft model only, direct generation, majority voting, and beam search. For majority voting and beam search, we employ a large number of samples, even exceeding the cost of using the target model alone, to approximate their converged performance.  
\textbf{(3) Speculative Decoding (SD).} We further consider speculative decoding, a method proposed for accelerating inference \citep{chen2023accelerating}.  
\textbf{(4) Reward-Guided Speculative Decoding (RSD).} RSD integrates a process reward model to score intermediate decoding steps and adaptively determine whether the target model should be invoked, thus striking a balance between efficiency and output quality \citep{liao2025reward}.  
\textbf{(5) Minimal Test-Time Intervention (MTI).} MTI is a training-free test-time intervention method that only applies additional guidance at a small subset of high-uncertainty (high-entropy) decoding positions, aiming to improve reasoning accuracy with minimal extra computation \citep{yang2025less}.

\textbf{Implement Details.}
All experiments are conducted on 8 NVIDIA A800 GPUs. For majority voting, beam search, and direct generation, we set the temperature to 0.7 and top$-p$ to 0.8, while the remaining methods are configured with temperature = 0 and top--$p$ = 1. Specifically, for RSD, we define a generation terminated by \textit{$\backslash$n$\backslash$n} as a reasoning step, which is then evaluated by a PRM with a threshold of 0.7. For MTI, we follow the original setup and conduct hyperparameter search over the intervention scale $\beta \in \{0.5, 1.0, 1.5, 2.0\}$, and report the best result for comparison. For EASD, to ensure a fair comparison under a comparable hyperparameter selection protocol, we first calculate the average entropy of the top 5\% highest-entropy tokens on the LIMO \citep{ye2025limo} dataset, and then map the resulting value to its nearest candidate in $\{0.5, 1.0, 1.5, 2.0\}$, which is used as the entropy threshold.

\begin{table*}[t]
\centering
\caption{Accuracy and decoding speed on benchmark datasets. TM denotes the target model, DM denotes the draft model, and PRM denotes the process reward model. TPS denotes the number of generated tokens per second.}
\scalebox{0.75}{
\begin{tabular}{lccccccccccc}
\toprule
Method & TM & DM & PRM & Olympiad & Minerva & Math500 & AMC23 & Aime24 & GPQA & Avg$\uparrow$&TPS$\uparrow$\\
\midrule
Single Model & 32B & - & - & 46.81 & 44.49 & 82.2 & 60 & 16.67& 45.96 & 49.35  &14.13\\
 Single Model& -& 7B& -& 35.70& 37.13& 73.2& 50& 10& 36.36&40.39 &36.75\\
Majority Voting (N=16) & - & 7B & - & 45.33& 43.01& 81& 67.5& 16.67& 41.41& 49.15 &36.75\\
Beam Search (N=16)& - & 7B & - & 40.15 & 39.34& 78& 62.5& 10& 42.42& 45.40  &28.53\\
SD & 32B & 7B & - & 45.33 & 45.22 & 80.8 & 57.5 & 13.33 & 49.49 & 48.61  &18.23\\
RSD \citep{liao2025reward} & 32B & 7B & 1.5B & 46.5& 46.32& \textbf{83.8}& \textbf{67.5}& 16.67& 50.50& 51.88 &16.89\\
MTI \citep{yang2025less} & 32B & - & - & 46.81& 44.85& 83& \textbf{67.5}& 16.67& 50& 51.47 &13.66\\
\midrule
EASD & 32B & 7B & - & \textbf{48.15} & \textbf{47.06} & 80.8 & 62.5 & \textbf{23.33}& \textbf{55.55} & \textbf{52.89}  &17.10\\
\bottomrule
\end{tabular}}

\scalebox{0.75}{
\begin{tabular}{lccccccccccc}
\toprule
Method & TM & DM & PRM & Olympiad & Minerva & Math500 & AMC23 & Aime24 & GPQA & Avg$\uparrow$&TPS$\uparrow$\\
\midrule
Single Model & 72B & - & - & 44.59 & 44.85 & \textbf{84} & 65 & 16.67& 46.96 & 50.34  &6.93\\
 Single Model& -& 7B& -& 35.70& 37.13& 73.2& 50& 10& 36.36&40.39 &36.75\\
Majority Voting (N=16) & - & 7B & - & 45.33& 43.01& 81& 67.5& 16.67& 41.41& 49.15 &36.75\\
Beam Search (N=16)& - & 7B & - & 40.15 & 39.34& 78& 62.5& 10& 42.42& 45.40  &28.53\\
SD & 72B & 7B & - & 44.15 & 44.12 & 83.8 & 60 & 13.33 & 50.50 & 49.31  &18.22\\
RSD \citep{liao2025reward}  & 72B & 7B & 1.5B &  45.62&  \textbf{46.32}&  83.8&  65& 16.67& 47.48& 50.81  &17.35\\
MTI \citep{yang2025less} & 72B & - & - & \textbf{45.77}& 44.49& 83.4& 62.5& \textbf{20}& \textbf{52.02}& 51.36 &6.81\\
\midrule
EASD & 72B & 7B & - & 44.74 & 44.85 & 83.6 & \textbf{67.5} & \textbf{20} &  \textbf{52.02} & \textbf{52.12}  &17.86\\
\bottomrule
    \end{tabular}}
\label{tab:experiment}
\end{table*}

\subsection{Main Results}
Table~\ref{tab:experiment} reports accuracy and decoding speed on six challenging reasoning benchmarks under two target-model scales. We evaluate standard single-model inference, draft-only test-time scaling methods, speculative decoding baselines, and guided or intervention-based approaches. Across both settings, \textbf{EASD consistently delivers the best overall accuracy while preserving the efficiency advantages of speculative decoding}. Three findings are most notable.

First, \textbf{EASD achieves the highest average accuracy in both regimes}. When using Qwen2.5-32B as the target model, EASD reaches an average accuracy of \textbf{52.89}. This improves over single-model 32B decoding by \textbf{3.54} percentage points and also surpasses the strongest draft-only scaling baseline, majority voting with $N=16$, by \textbf{3.74} points. When scaling the target model to Qwen2.5-72B, EASD remains the top method with \textbf{52.12} average accuracy, exceeding single-model 72B decoding by \textbf{1.78} points. The fact that EASD ranks first under both target sizes indicates that the improvement is not an artifact of a particular model scale, but rather reflects a systematic benefit from entropy-aware token rejection during decoding.

Second, \textbf{EASD outperforms guided speculative decoding and sparse intervention baselines without relying on auxiliary models}. In particular, EASD consistently improves upon \textbf{RSD}~\citep{liao2025reward}, even though RSD introduces a process reward model to score intermediate steps. EASD exceeds RSD by \textbf{1.01} points in the 32B setting and by \textbf{1.31} points in the 72B setting. EASD also outperforms \textbf{MTI}~\citep{yang2025less} in both regimes, with gains of \textbf{1.42} points for the 32B target and \textbf{0.76} points for the 72B target. These comparisons suggest that token-level uncertainty gating can be more effective than reward-guided verification or occasional intervention, while keeping the overall pipeline lightweight and training-free.

Third, \textbf{EASD provides a stronger accuracy--throughput trade-off than RSD and MTI, achieving both higher speed and higher accuracy}. In the 32B regime, EASD runs at \textbf{17.10} tokens per second, which is faster than RSD at \textbf{16.89} tokens per second and substantially faster than MTI at \textbf{13.66} tokens per second. Importantly, this speed advantage does not come at the expense of quality, since EASD also attains the best average accuracy among all methods in this regime. The same pattern holds in the 72B regime. EASD achieves \textbf{17.86} tokens per second, exceeding RSD at \textbf{17.35} tokens per second and far surpassing MTI at \textbf{6.81} tokens per second, while still achieving the highest average accuracy overall. This indicates that EASD preserves most of the efficiency benefit of speculative decoding, and that its selective rejection-and-resampling mechanism improves reliability with minimal throughput loss.

Overall, these results show that EASD is not only a consistent accuracy improvement over standard speculative decoding, but also a practical alternative to reward-guided and intervention-based test-time methods. It delivers the best average accuracy across both target-model scales and simultaneously maintains strong decoding speed, making it an appealing drop-in decoding strategy for reasoning-focused workloads.

\subsection{Ablation Experiments}
To quantify the impact of each design choice in \textbf{EASD}, we conduct ablation experiments by selectively removing the two core signals used to trigger entropy-aware rejection, namely the \emph{draft-model entropy} and the \emph{top-$n$ overlap regulation}. Concretely, we evaluate four variants: \textbf{(1) EASD}, the full method; \textbf{(2) EASD w/o Overlap}, which disables the overlap regulation so the decision no longer depends on token-level agreement between the draft and target distributions; \textbf{(3) EASD w/o $H_d$}, which removes the draft-model entropy signal and relies only on the target model's uncertainty; and \textbf{(4) EASD w/o ($H_d$ + Overlap)}, which removes both signals and thus yields the most simplified gating rule. These variants allow us to isolate whether EASD's improvements primarily come from incorporating the draft model's uncertainty, from explicitly modeling draft--target agreement, or from their interaction.

Table~\ref{tab:addlabel} shows that \textbf{the full EASD configuration is the most reliable across benchmarks and target scales}. For both the 32B and 72B target models, EASD attains the highest average accuracy, indicating that the combination of draft entropy and overlap regulation produces the most robust rejection decisions. This trend supports the central motivation of EASD: a token should be reconsidered not only when the target model is uncertain, but also when the draft model is uncertain in a compatible way, since such cases often correspond to shared confusion where speculative decoding may otherwise lock in an error.

\textbf{Draft-model entropy is the dominant contributor to accuracy gains.} Removing $H_d$ causes the largest average degradation in both regimes. With the 32B target, average accuracy drops from \textbf{52.89} to \textbf{50.85}, and with the 72B target it drops from \textbf{52.12} to \textbf{50.86}. The declines are particularly visible on the reasoning-intensive datasets, suggesting that $H_d$ provides a crucial early-warning signal about low-confidence draft proposals. Intuitively, even if the target model's entropy is moderately high, the draft model may reveal additional uncertainty about the same decision boundary, and this joint uncertainty is exactly where resampling is most beneficial. Without $H_d$, the method degenerates toward a target-only uncertainty rule, which appears insufficient to consistently detect and correct speculative errors.

\textbf{Overlap regulation further improves stability and complements entropy-based gating.} Removing overlap typically reduces average accuracy, although the effect is more moderate than removing $H_d$. In the 32B regime, the average decreases from \textbf{52.89} to \textbf{51.77}, and in the 72B regime it decreases from \textbf{52.12} to \textbf{51.93}. This behavior is consistent with the role of overlap as a disambiguation signal: high overlap indicates that the two models are concentrating probability mass on a similar candidate set, so an aligned token is more likely to reflect a shared but potentially brittle decision. By conditioning rejection on overlap, EASD avoids unnecessary resampling when the draft and target disagree sharply, a scenario where forcing rejection can be either redundant or harmful. The overlap term therefore acts as a stabilizer that makes entropy-based rejection more selective.

\textbf{Removing both signals produces the weakest variant, confirming their complementarity.} When both $H_d$ and overlap regulation are disabled, the average accuracy drops further to \textbf{49.36} in the 32B setting and to \textbf{49.53} in the 72B setting, yielding the largest overall degradation among the variants. Beyond lowering the average, this variant is also less consistent across datasets, with notable drops on benchmarks that require long-horizon reasoning. This suggests that draft entropy and overlap regulation play complementary roles: $H_d$ supplies a strong uncertainty cue, while overlap constrains the cue to cases where the two distributions exhibit meaningful agreement, producing more reliable intervention points.

Overall, the ablation study confirms that \textbf{EASD's gains are not attributable to a single heuristic}, but rather to the joint use of (i) uncertainty from the draft model and (ii) an agreement-based filter between the two models. Together, these components yield more accurate and better-calibrated rejection decisions, which in turn translate into consistent improvements over simplified variants across both target-model scales.

\begin{table*}[t]
\centering
\caption{Ablation Study on the Effects of Draft Model Entropy and Overlap Regulation.}
\scalebox{0.9}{
\begin{tabular}{lccccccccc}
\toprule
Method & TM & DM & Olympiad & Minerva & Math500 & AMC23 & Aime24 & GPQA & Avg \\
\midrule
EASD & 32B & 7B &  \textbf{48.15} & \textbf{47.06} & \textbf{80.8}& 62.5 & \textbf{23.33}& \textbf{55.55} & \textbf{52.89} \\
~~~~w/o Overlap & 32B & 7B &  47.41& 44.85& 80.8& \textbf{65}& 20& 52.53& 51.77\\
~~~~w/o $H_d$ & 32B & 7B &  46.52& 44.85& 80.2& 62.5& 20& 51.01& 50.85\\
~~~~w/o ($H_d$ + Overlap) & 32B & 7B &  47.41& 46.32& 80.6& 60& 13.33& 48.48& 49.36\\
\bottomrule
\end{tabular}}

\scalebox{0.9}{
\begin{tabular}{lccccccccc}
\toprule
Method & TM & DM & Olympiad & Minerva & Math500 & AMC23 & Aime24 & GPQA & Avg \\
\midrule
EASD & 72B & 7B  & \textbf{44.74}& \textbf{44.85}& \textbf{83.6}& 67.5& \textbf{20}&  52.02& \textbf{52.12} \\
~~~~w/o Overlap & 72B & 7B &  44.59& 44.49& 82.8& \textbf{70}& 16.67& \textbf{53.03}& 51.93\\
~~~~w/o $H_d$ & 72B & 7B &  43.85& 44.12& 83.0& 67.5& 16.67& 50& 50.86\\
~~~~w/o ($H_d$ + Overlap) & 72B & 7B &  44.15& 44.49& 82.4& 62.5& 16.67& 46.97& 49.53\\
\bottomrule
\end{tabular}}
\label{tab:addlabel}
\end{table*}

\subsection{Hyperparameter Sensitivity Study}
\begin{figure}[t]
  \centering
  \begin{subfigure}[t]{0.49\columnwidth}
    \centering
    \includegraphics[width=\linewidth,height=0.22\textheight,keepaspectratio]{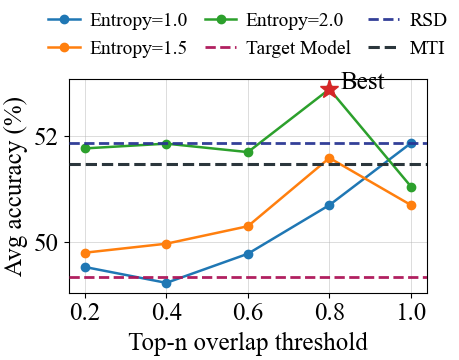}
  \end{subfigure}\hfill
  \begin{subfigure}[t]{0.49\columnwidth}
    \centering
    \includegraphics[width=\linewidth,height=0.22\textheight,keepaspectratio]{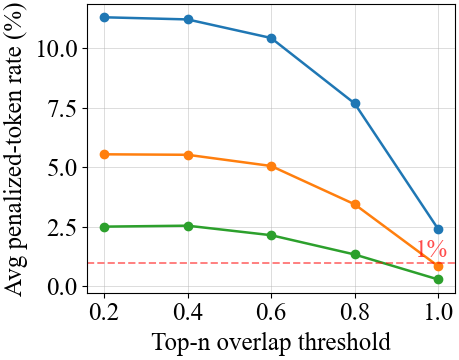}
  \end{subfigure}
  \caption{EASD performance under different entropy and top-$n$ threshold settings (\%): the left panel shows accuracy, and the right panel shows the penalized-token rate.}
  \label{fig:accuracy}
\end{figure}

We conduct a hyperparameter sensitivity study to characterize how the entropy threshold $\tau_H$ and the top-$n$ overlap threshold $\tau_O$ shape the behavior of EASD. These thresholds jointly gate the penalty mechanism, and therefore directly control (a) how often EASD intervenes and (b) how much those interventions translate into accuracy gains. We sweep $\tau_H \in {1, 1.5, 2}$ and $\tau_O \in {0.2, 0.4, 0.6, 0.8, 1}$, and report in a single figure (Figure \ref{fig:accuracy}) both (i) average accuracy over six reasoning benchmarks and (ii) intervention strength measured by the penalized-token frequency.

The intervention rate varies monotonically with both thresholds. Because penalization is activated only when \emph{both} conditions are satisfied (high entropy \emph{and} high top-$n$ overlap), increasing either $\tau_H$ or $\tau_O$ makes the trigger condition stricter and reduces the fraction of penalized tokens. The penalized-token frequency panel in Fig.~\ref{fig:accuracy} therefore forms a smooth, monotone surface over the grid: smaller thresholds correspond to more frequent interventions, while larger thresholds yield increasingly sparse interventions. This behavior is important in practice because it means $\tau_H$ and $\tau_O$ act as predictable dials for intervention sparsity, rather than inducing qualitatively different decoding regimes or unstable dynamics.

Accuracy remains robust across a wide portion of the grid. The accuracy panel in Fig.~\ref{fig:accuracy} shows that EASD outperforms the target-only baseline for essentially all configurations, with only a single degraded point at $(\tau_H=1,\tau_O=0.4)$. More broadly, accuracy changes smoothly as thresholds vary: we do not observe sharp performance cliffs, which indicates EASD does not rely on a narrowly tuned operating point. Notably, multiple threshold pairs reach or exceed the accuracy of RSD and MTI while using no process reward model, suggesting that the entropy-and-overlap gate can provide competitive test-time guidance without auxiliary supervision. 

Reading the two panels together highlights a clear selectivity principle: the best results arise when intervention is \emph{rare but well-targeted}. In the high-intervention regime (penalized-token frequency above roughly 1\%), tightening either threshold typically reduces the penalization rate and improves accuracy. This pattern suggests that overly frequent rejection can dilute the effectiveness of resampling, likely because it perturbs many steps that are uncertain yet not truly error-inducing, thereby injecting unnecessary stochasticity. By contrast, as the thresholds increase and intervention becomes sparse, accuracy remains consistently strong and often near its peak.

These results indicate that EASD’s gains are driven by correcting a small subset of high-uncertainty, high-overlap steps rather than by pervasive rejection. In other words, the overlap condition is not merely permissive; it helps concentrate interventions on "suspicious agreement" events where both models are uncertain but still propose highly similar top candidates, a signature that can coincide with systematic mistakes shared across related models.

Overall, the sweep identifies a broad robust operating region where EASD achieves strong accuracy with low penalized-token frequency, supporting the practicality of threshold selection.

\section{Conclusion}

In this work, we propose Entropy-Aware Speculative Decoding (EASD), a training-free extension of speculative decoding that leverages entropy signals from both draft and target models to prevent low-confidence tokens from propagating errors. EASD retains the efficiency advantages of standard speculative decoding while achieving substantial improvements in token-level accuracy across challenging reasoning benchmarks. Our results show that EASD outperforms recent speculative decoding variants and can even surpass the inherent performance of the target model in certain settings. These findings suggest that speculative decoding can provide useful uncertainty signals beyond acceleration. Overall, EASD offers a lightweight, effective, and broadly applicable enhancement to efficient LLM inference, and can be integrated into existing speculative decoding pipelines with minimal implementation changes.

\section{Limitations}

Although EASD shows consistent improvements over recent speculative decoding methods, our study has several limitations. First, the evaluation has not yet been extended to broader real-world reasoning scenarios, which may involve more diverse tasks, domains, and deployment constraints. Second, our experiments primarily focus on medium- to large-scale LLMs, leaving the effectiveness of EASD on smaller, larger, and domain-specific models for future investigation. Finally, while EASD is training-free, its interaction with advanced alignment techniques and domain-adapted fine-tuning remains underexplored. Future work may further investigate adaptive threshold selection to improve robustness across different tasks and model settings.

\section*{Acknowledgements}
This work is supported by the National Natural Science Foundation of China (72204087), the Shanghai Planning Office of Philosophy and Social Science Youth Project (2022ETQ001), the Chenguang Program of Shanghai Education Development Foundation and Shanghai Municipal Education Commission (23CGA28), the Shanghai Pujiang Program (23PJC030), Young Elite Scientists Sponsorship Program by CAST (YESS20240562), and the Fundamental Research Funds for the Central Universities, China. We also appreciate the constructive comments from the anonymous reviewers.

\bibliography{custom}

\appendix

\newpage

\section{Pseudocode for Entropy-Aware Speculative Decoding (EASD)}
\label{appendix:EASD}

The following pseudocode summarizes the workflow of \textbf{Entropy-Aware Speculative Decoding (EASD)}, an extension of standard speculative decoding (SD).  
EASD retains the core two-model structure of SD, where a draft model proposes candidate tokens and a target model verifies them. 
The key innovation of EASD is the \textit{dynamic penalty} mechanism, which is applied when both models exhibit high entropy and their top-$n$ token predictions significantly overlap.  
This mechanism suppresses the draft model's top token and renormalizes the target distribution, encouraging alternative selections and reducing error propagation.

\begin{center}
\begin{minipage}{0.98\columnwidth}
\small
\hrule
\vspace{2pt}
\refstepcounter{algorithm}
\noindent\textbf{Algorithm~\thealgorithm\ Entropy-Aware Speculative Decoding (EASD)}
\label{alg:EASD}
\vspace{2pt}
\hrule
\vspace{4pt}

\begin{algorithmic}[1]
\REQUIRE Context $s$; target model $\mathrm{LLM}_{\mathrm{targ}}$;
draft model $\mathrm{LLM}_{\mathrm{draft}}$
\REQUIRE Draft length $M$; top-$n$ size $n$; thresholds $\tau_H,\tau_O$
\ENSURE Generated sequence $y$

\STATE $y\gets []$
\WHILE{$\mathrm{EOS}\notin y$}
    \STATE $x\gets s\oplus y$
    \STATE $(c_i,P_d^{(i)})_{i=1}^{M}
    \gets\mathrm{LLM}_{\mathrm{draft}}^{(M)}(x)$
    \STATE $(P_t^{(i)})_{i=1}^{M+1}
    \gets\mathrm{LLM}_{\mathrm{targ}}(c_1,\ldots,c_M;x)$
    \STATE $\textit{allAccepted}\gets\textbf{true}$

    \FOR{$i=1$ \TO $M$}
        \STATE Compute $H_d^{(i)}$, $H_t^{(i)}$, and
        $O^{(i)}=|T_{d,i}^{n}\cap T_{t,i}^{n}|/n$

        \IF{$H_d^{(i)}>\tau_H$ \textbf{and}
        $H_t^{(i)}>\tau_H$ \textbf{and} $O^{(i)}\ge\tau_O$}
            \STATE $\widetilde{P}_t^{(i)}(c_i)\gets 0$
            \STATE $\widetilde{P}_t^{(i)}(v)
            \gets \dfrac{P_t^{(i)}(v)}
            {1-P_t^{(i)}(c_i)},\quad \forall v\neq c_i$
            \STATE $r_i\gets\operatorname{Decode}(\widetilde{P}_t^{(i)})$
            \STATE $y\gets y\oplus r_i$;
            $\textit{allAccepted}\gets\textbf{false}$
            \STATE \textbf{break}
        \ELSIF{$\operatorname{Verify}(c_i,P_d^{(i)},P_t^{(i)})$}
            \STATE $y\gets y\oplus c_i$
        \ELSE
            \STATE $r_i\gets\operatorname{Decode}(P_t^{(i)})$
            \STATE $y\gets y\oplus r_i$;
            $\textit{allAccepted}\gets\textbf{false}$
            \STATE \textbf{break}
        \ENDIF
    \ENDFOR

    \IF{\textit{allAccepted}}
        \STATE $y\gets y\oplus
        \operatorname{Decode}(P_t^{(M+1)})$
    \ENDIF
\ENDWHILE
\RETURN $y$
\end{algorithmic}

\vspace{2pt}
\hrule
\end{minipage}
\end{center}

\section{Generalization to Non-Reasoning Generation Tasks}
\label{appendix:non_reasoning}

Although EASD is mainly motivated by reasoning-oriented generation, it is also important to examine whether the proposed entropy-aware rejection mechanism introduces undesirable side effects on more general text generation tasks. Reasoning benchmarks often involve highly structured multi-step inference, where an early uncertain token may substantially change the subsequent reasoning trajectory. In contrast, non-reasoning generation tasks such as summarization and translation usually allow a wider range of acceptable token choices, as long as the generated output preserves the source information and remains fluent. Therefore, these tasks provide a useful testbed for evaluating the stability and generality of EASD beyond mathematical or scientific reasoning scenarios.

To this end, we evaluate EASD on two representative non-reasoning generation tasks: CNN/DailyMail summarization and IWSLT translation. CNN/DailyMail is used to assess the quality of long-form abstractive summarization, where the model is expected to preserve salient information from the source document. IWSLT translation is used to evaluate sequence-to-sequence generation under a different linguistic transformation setting. For summarization, we report ROUGE scores, including CNN R-1, R-2, and R-3. For translation, we report BLEU. We compare EASD with standalone target-model decoding and standard speculative decoding (SD) under the same target model setting.

Table~\ref{tab:non_reasoning} presents the results for Qwen-2.5 (7B--32B) and Qwen-2.5 (7B--72B). Overall, EASD remains stable and performs comparably to both the target model and standard SD on non-reasoning tasks. On CNN/DailyMail summarization, EASD achieves ROUGE scores that are very close to the corresponding target-model and SD baselines. For Qwen-2.5 (7B--32B), EASD obtains 33.68 on CNN R-1, 11.24 on CNN R-2, and 4.54 on CNN R-3, which are comparable to the target and SD results. For Qwen-2.5 (7B--72B), EASD achieves the highest CNN R-1 score among the three decoding methods, while maintaining similar R-2 and R-3 scores. These results indicate that entropy-aware token rejection does not substantially disrupt summarization quality.

On IWSLT translation, EASD also maintains comparable BLEU scores. With Qwen-2.5 (7B--32B), EASD slightly improves BLEU over both the standalone target model and SD. With Qwen-2.5 (7B--72B), the BLEU scores of the three methods remain very close. This suggests that EASD can preserve translation quality even though its intervention mechanism is designed around token-level uncertainty rather than task-specific translation objectives.

These findings provide two useful observations. First, EASD is not narrowly restricted to reasoning benchmarks: when applied to non-reasoning generation tasks, it generally maintains output quality rather than causing systematic degradation. Second, the improvements on summarization and translation are relatively modest compared with those observed on reasoning benchmarks. This pattern is consistent with the design motivation of EASD. Entropy-aware rejection is most beneficial when an uncertain token can strongly affect the subsequent reasoning path, which is more common in mathematical and scientific reasoning tasks. For summarization and translation, where several alternative continuations may be similarly acceptable, EASD mainly acts as a stable decoding modification rather than a strong performance-enhancing mechanism.

Overall, the non-reasoning results suggest that EASD provides a favorable trade-off: it improves reasoning-oriented generation in settings where token-level uncertainty is highly consequential, while remaining stable on broader generation tasks such as summarization and translation.

\begin{table}[!htbp]
  \centering
  \scriptsize
  \setlength{\tabcolsep}{3pt}
  \renewcommand{\arraystretch}{0.95}
  \caption{Generalization results on non-reasoning generation tasks. CNN R-1, R-2, and R-3 denote ROUGE scores on CNN/DailyMail summarization, while IWSLT BLEU reports translation performance.}
  \label{tab:non_reasoning}
  \begin{tabular}{lrrrr}
    \toprule
    Method & CNN R-1 & CNN R-2 & CNN R-3 & IWSLT BLEU \\
    \midrule
    \multicolumn{5}{c}{\textbf{Qwen-2.5 (7B--32B)}} \\
    Target & \textbf{33.72} & 11.31 & 4.59 & 30.81 \\
    SD     & 33.65 & \textbf{11.34} & \textbf{4.65} & 30.93 \\
    EASD   & 33.68 & 11.24 & 4.54 & \textbf{30.99} \\
    \midrule
    \multicolumn{5}{c}{\textbf{Qwen-2.5 (7B--72B)}} \\
    Target & 34.71 & \textbf{12.05} & \textbf{5.03} & \textbf{32.89} \\
    SD     & 34.71 & \textbf{12.05} & \textbf{5.03} & 32.87 \\
    EASD   & \textbf{34.75} & 12.00 & 4.99 & 32.88 \\
    \bottomrule
  \end{tabular}
\end{table}

\section{Safety on Code and Structured Outputs}
\label{appendix:structured_outputs}

The entropy-aware intervention in EASD deliberately suppresses an uncertain agreed token and redirects generation toward an alternative continuation. While this behavior is useful for exploring different reasoning trajectories, it is important to verify that it does not introduce additional degradation on syntax-sensitive tasks, where even a small token-level change may invalidate the entire output. We therefore evaluate EASD on LiveCodeBench-V6 for code generation and StructEval-T for structured generation. StructEval-T covers five commonly used formats: JSON, XML, YAML, CSV, and TOML. We compare EASD with standalone target-model decoding and standard speculative decoding (SD) using two Qwen-2.5 draft--target pairs.

As shown in Table~\ref{tab:structured_outputs}, EASD does not cause additional degradation relative to SD. With the 7B--32B pair, EASD matches SD on both LiveCodeBench-V6 and StructEval-T. With the 7B--72B pair, EASD improves over SD by 1.14 points on LiveCodeBench-V6 and by 0.41 points on StructEval-T. It also exceeds standalone target decoding by 0.57 points on LiveCodeBench-V6. These results indicate that the selective intervention used by EASD remains safe on outputs governed by strict syntactic and structural constraints.

\begin{table}[!htbp]
  \centering
  \scriptsize
  \renewcommand{\arraystretch}{0.95}
  \caption{Results on syntax-sensitive generation tasks. LiveCodeBench-V6 evaluates code generation, while StructEval-T evaluates structured outputs in JSON, XML, YAML, CSV, and TOML.}
  \label{tab:structured_outputs}
  \setlength{\tabcolsep}{4pt}
  \begin{tabular}{lrr}
    \toprule
    Method & LiveCodeBench-V6 & StructEval-T \\
    \midrule
    \multicolumn{3}{c}{\textbf{Qwen-2.5 (7B--32B)}} \\
    Target & \textbf{17.71} & \textbf{78.74} \\
    SD     & 17.14 & 77.46 \\
    EASD   & 17.14 & 77.46 \\
    \midrule
    \multicolumn{3}{c}{\textbf{Qwen-2.5 (7B--72B)}} \\
    Target & 20.00 & \textbf{84.79} \\
    SD     & 19.43 & 83.37 \\
    EASD   & \textbf{20.57} & 83.78 \\
    \bottomrule
  \end{tabular}
\end{table}

\section{Robustness under Alternative Decoding Settings}
\label{appendix:decoding_robustness}

The main experiments compare decoding methods under a fixed inference configuration. We conduct two additional evaluations to determine whether the improvements of EASD can be explained by stochastic variation or by the use of a particular decoding paradigm. First, we compare EASD against repeated target-only sampling on all six original reasoning benchmarks. Second, we evaluate EASD under greedy decoding, independent sampling, and majority voting on AIME25.

\subsection{Comparison with Repeated Target-Only Sampling}
\label{appendix:repeated_sampling}

We sample from the standalone target model at temperature 0.7 using three independent runs per prompt. Table~\ref{tab:repeated_sampling} reports both the mean of the three runs and an optimistic per-dataset best score obtained by selecting the strongest of the three runs separately for each benchmark. The latter is included as a deliberately favorable reference for target-only sampling rather than as a deployable single-run configuration.

For Qwen-2.5 (7B--32B), EASD achieves an average score of 52.89, exceeding both the three-run sampling mean of 49.20 and the per-dataset best score of 51.29. The same overall pattern holds for Qwen-2.5 (7B--72B): EASD achieves 52.12, compared with 50.01 for the sampling mean and 51.88 for the per-dataset best result. Thus, the average improvement of EASD is not reproduced by repeated target-only sampling, even when the target baseline is allowed to select its best run independently on each dataset.

\begin{table}[!htbp]
  \centering
  \scriptsize
  \renewcommand{\arraystretch}{0.95}
  \caption{Comparison with target-only sampling at temperature 0.7 using three independent runs per prompt. ``Per-dataset best'' selects the best of the three runs separately for each benchmark and therefore serves as an optimistic reference.}
  \label{tab:repeated_sampling}
  \setlength{\tabcolsep}{1.2pt}
  \begin{tabular}{lrrrrrrr}
    \toprule
    Setting & Oly. & Min. & M500 & AMC & AIME & GPQA & Avg. \\
    \midrule
    \multicolumn{8}{c}{\textbf{Qwen-2.5 (7B--32B)}} \\
    3-run mean       & 44.84 & 44.49 & 80.60 & 60.00 & 12.22 & 53.03 & 49.20 \\
    Per-dataset best & 45.93 & 45.59 & \textbf{81.00} & \textbf{65.00} & 16.67 & 53.54 & 51.29 \\
    EASD             & \textbf{48.15} & \textbf{47.06} & 80.80 & 62.50 & \textbf{23.33} & \textbf{55.55} & \textbf{52.89} \\
    \midrule
    \multicolumn{8}{c}{\textbf{Qwen-2.5 (7B--72B)}} \\
    3-run mean       & 45.33 & 42.40 & 82.73 & 60.83 & 17.78 & 51.01 & 50.01 \\
    Per-dataset best & \textbf{45.63} & 43.38 & 83.20 & 65.00 & \textbf{20.00} & \textbf{54.04} & 51.88 \\
    EASD             & 44.74 & \textbf{44.85} & \textbf{83.60} & \textbf{67.50} & \textbf{20.00} & 52.02 & \textbf{52.12} \\
    \bottomrule
  \end{tabular}
\end{table}

\subsection{Evaluation under Greedy Decoding, Sampling, and Majority Voting}
\label{appendix:aime25_decoding}

We further evaluate all 30 AIME25 problems under three inference paradigms. Greedy decoding evaluates whether EASD can improve a single deterministic reasoning trajectory. For sampling, we generate eight responses per problem and report the mean accuracy and standard deviation across the eight responses. For majority voting, we aggregate the same eight responses and evaluate the resulting voted answer. This evaluation includes both Qwen-2.5 model pairs and a DeepSeek-R1 1.5B--7B pair, allowing us to test robustness across decoding settings and applicability to a modern reasoning-model family.

Table~\ref{tab:aime25_decoding} shows that EASD consistently improves over standalone target decoding in all nine comparisons. For Qwen-2.5 (7B--32B), EASD improves majority-voting accuracy from 16.67 to 23.33, average sampling accuracy from $12.50 \pm 3.45$ to $14.17 \pm 4.63$, and greedy accuracy from 16.67 to 23.33. For Qwen-2.5 (7B--72B), the corresponding improvements are from 16.67 to 20.00 under majority voting, from $12.08 \pm 3.54$ to $14.58 \pm 4.69$ under sampling, and from 13.33 to 20.00 under greedy decoding. These consistent gains provide evidence that the improvements of EASD are not tied to a particular decoding paradigm. In particular, the greedy results show that EASD can redirect an individual reasoning trajectory, while the sampling results indicate an improvement in the overall quality of alternative trajectories. The majority-voting results further demonstrate that these improvements remain useful after answer aggregation.

The DeepSeek-R1 results provide an additional evaluation on models explicitly optimized for reasoning. Using a 1.5B draft model and a 7B target model, EASD improves majority-voting accuracy from 40.00 to 56.67, average sampling accuracy from $32.50 \pm 4.96$ to $40.83 \pm 4.27$, and greedy accuracy from 23.33 to 33.33. The improvements across all three inference paradigms demonstrate that EASD is also applicable to a modern reasoning-model family and is not limited to conventional instruction-tuned models. Overall, the results across the three model pairs show that EASD consistently improves reasoning accuracy under deterministic generation, stochastic sampling, and answer aggregation.

\begin{table}[!htbp]
  \centering
  \scriptsize
  \setlength{\tabcolsep}{3pt}
  \renewcommand{\arraystretch}{0.95}
  \caption{Accuracy on all 30 AIME25 problems under greedy decoding,
  sampling with eight responses per problem, and majority voting over
  eight responses (\%). Sampling results are reported as mean
  $\pm$ standard deviation across the eight responses.}
  \label{tab:aime25_decoding}
  \begin{tabular}{lccc}
    \toprule
    \multicolumn{1}{c}{Method}
      & Majority Vote & Sampling & Greedy \\
    \midrule

    \multicolumn{4}{c}{\textbf{Qwen-2.5 (7B--32B)}} \\
    Target & 16.67 & $12.50 \pm 3.45$ & 16.67 \\
    EASD   & \textbf{23.33}
           & $\mathbf{14.17} \boldsymbol{\pm} \mathbf{4.63}$
           & \textbf{23.33} \\

    \midrule
    \multicolumn{4}{c}{\textbf{Qwen-2.5 (7B--72B)}} \\
    Target & 16.67 & $12.08 \pm 3.54$ & 13.33 \\
    EASD   & \textbf{20.00}
           & $\mathbf{14.58} \boldsymbol{\pm} \mathbf{4.69}$
           & \textbf{20.00} \\

    \midrule
    \multicolumn{4}{c}{\textbf{DeepSeek-R1 (1.5B--7B)}} \\
    Target & 40.00 & $32.50 \pm 4.96$ & 23.33 \\
    EASD   & \textbf{56.67}
           & $\mathbf{40.83} \boldsymbol{\pm} \mathbf{4.27}$
           & \textbf{33.33} \\
    \bottomrule
  \end{tabular}
\end{table}

\section{Empirical Analysis of High-Entropy Token Distribution}
\label{appendix:entropy_distribution}

To further understand how EASD affects uncertain token decisions, we compare the distributions of high-entropy tokens before and after applying EASD using the Qwen2.5 model pairs. Table~\ref{tab:entropy-distribution} reports the most frequent tokens occurring at high-entropy positions under standalone target decoding and EASD. The original distribution is relatively dispersed. Among the identified tokens, \textit{``Given''} has the highest proportion at 2.34\%, followed by \textit{``will''} and \textit{``Let''} at 1.55\%, \textit{``need''} at 1.53\%, and \textit{``Thus''} at 1.50\%. Other tokens, including \textit{``calculate''}, \textit{``Now''}, \textit{``consider''}, and \textit{``determine''}, occur with lower proportions. Overall, these tokens cover multiple stages of mathematical reasoning, including the introduction of premises, formulation of assumptions, intermediate calculations, and derivation of conclusions.

After applying EASD, the composition of high-entropy tokens changes noticeably. The proportion of \textit{``Given''} increases from 2.34\% to 2.64\%, while \textit{``Thus''} rises from 1.50\% to 2.27\% and \textit{``consider''} increases from 1.01\% to 1.77\%. These changes indicate that high-entropy positions under EASD become more concentrated around premise identification, analytical transitions, and conclusion formation. In contrast, \textit{``will''} decreases from 1.55\% to 1.00\%, and \textit{``Let''} decreases from 1.55\% to 1.33\%. Tokens such as \textit{``need''}, \textit{``Now''}, and \textit{``determine''} no longer appear among the most frequent high-entropy tokens, while \textit{``understand''}, \textit{``Using''}, and \textit{``use''} emerge after applying EASD. This replacement pattern further suggests that EASD shifts uncertain positions from generic planning expressions toward tokens that more explicitly organize reasoning operations.

Overall, EASD systematically redistributes high-entropy token occurrences rather than merely reducing or uniformly shifting token uncertainty. The increased presence of tokens associated with premises, analytical transitions, and conclusions is consistent with the intended role of EASD: intervening at uncertain reasoning positions and redirecting generation toward more structured reasoning decisions. Together with the consistent accuracy improvements under repeated sampling, greedy decoding, and majority voting, this analysis provides additional empirical evidence that EASD affects the reasoning process rather than obtaining improvements solely from stochastic variation.

\begin{table}[!htbp]
  \centering
  \scriptsize
  \setlength{\tabcolsep}{2.5pt}
  \renewcommand{\arraystretch}{0.95}
  \caption{Distribution of frequently occurring high-entropy tokens before and after applying EASD.}
  \label{tab:entropy-distribution}
  \begin{tabular}{lccc}
    \toprule
    Token & Original (\%) & EASD (\%) & Change \\
    \midrule
    Given
      & 2.34 & 2.64 & $\uparrow$ Premise focus \\
    Thus
      & 1.50 & 2.27 & $\uparrow$ Conclusion focus \\
    consider
      & 1.01 & 1.77 & $\uparrow$ Analytical reasoning \\
    Let
      & 1.55 & 1.33 & $\downarrow$ Hypothetical usage \\
    will
      & 1.55 & 1.00 & $\downarrow$ Speculative usage \\
    need
      & 1.53 & -- & Removed \\
    Now
      & 1.08 & -- & Removed \\
    determine
      & 0.87 & -- & Removed \\
    understand
      & -- & 0.97 & Newly introduced \\
    Using
      & -- & 0.92 & Method application \\
    use
      & -- & 0.91 & Method application \\
    \bottomrule
  \end{tabular}
\end{table}

\begin{figure*}[!t]
  \centering
  \includegraphics[width=0.90\textwidth]{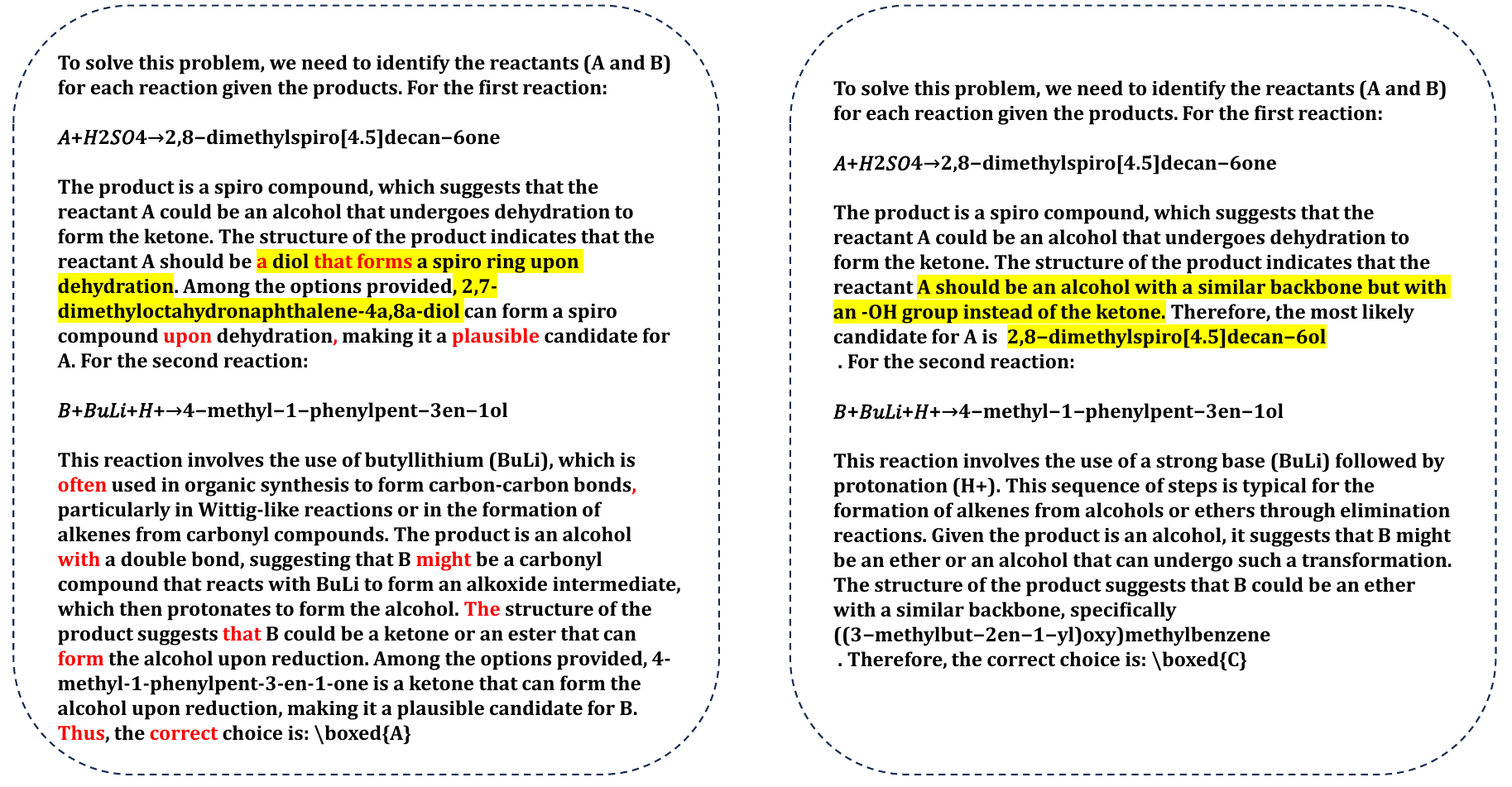}
  \caption{\textbf{Success Case:} Answer comparison on a GPQA reasoning problem, illustrating the difference between the original output and the EASD-enhanced output.}
  \label{fig:GPQA}
\end{figure*}

\section{Case Study}

\subsection{Success Case}

The discrepancy between the original answer and the EASD-enhanced answer can be traced back to the first decisive token used to identify the initial reactant. For the first reaction,
\[
A + H_2SO_4 \rightarrow 2,8\text{-dimethylspiro}[4.5]\text{decan}-6\text{one},
\]
the original answer begins with the phrase \texttt{an alcohol}, whereas the EASD-enhanced answer starts with \texttt{a diol}. Although this difference may appear minor at the token level, it substantially changes the chemical hypothesis adopted by the model. In particular, the choice between \texttt{an} and \texttt{a} constrains the noun that follows and therefore commits the generation process to different classes of candidate reactants.

In the original answer, the phrase \texttt{an alcohol} frames the subsequent reasoning around a monohydroxyl precursor. Once this assumption is introduced, the model begins to explore compounds containing only a single hydroxyl group and attempts to explain the formation of the target product through transformations of such structures. However, these candidates do not naturally support the dehydration and cyclization pathway required to construct the spirocyclic ketone. Consequently, the initial incorrect assumption narrows the search space in an unfavorable direction and causes the following reasoning steps to remain internally consistent but chemically inappropriate.

By contrast, the EASD-enhanced generation begins with \texttt{a diol}, emphasizing a precursor containing two hydroxyl groups. This provides a more suitable structural basis for the subsequent acid-catalyzed transformation. Under the action of sulfuric acid, an appropriately positioned diol can undergo dehydration and rearrangement or cyclization, thereby enabling the formation of the required spiro ring and ketone functionality. The revised token therefore directs the model toward a class of reactants that is substantially more compatible with the structural characteristics of the target product.

In summary, the success of EASD in this case can be attributed to generating the more appropriate initial phrase, \texttt{a diol}. This modification not only guides the model toward chemically viable candidates but also prevents the initial misconception from influencing the remainder of the reasoning process. The case thus provides a concrete illustration of how token-level intervention at an early and uncertain decoding position can produce a meaningful improvement in the final answer.

\begin{figure*}[!t]
  \centering
  \includegraphics[width=0.90\textwidth]{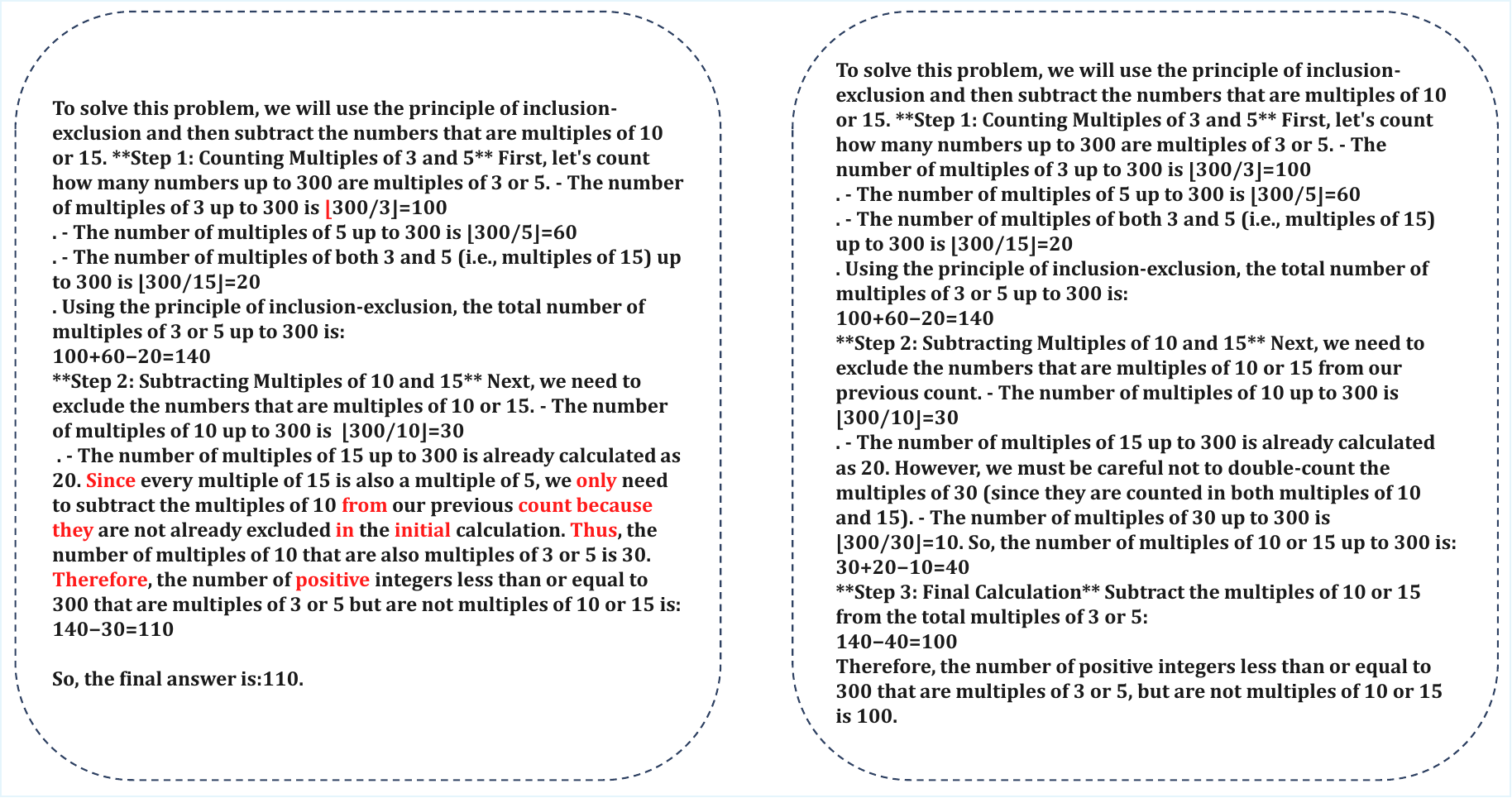}
  \caption{\textbf{Bad Case:} Answer comparison on a Math reasoning problem, illustrating the difference between the original output and the EASD-output.}
  \label{fig:bad_case}
\end{figure*}

\subsection{Bad Case}

Nevertheless, EASD does not improve every generation. In this bad case, the EASD-enhanced model produced an incorrect answer because of subtle interactions between several locally generated tokens. In particular, the consecutive appearance of the tokens \texttt{since} and \texttt{only} altered the logical structure of the explanation and directed the model toward an incomplete counting procedure. Although neither token is inherently problematic, their combination encouraged the model to introduce an overly restrictive justification for which elements should be excluded from the count.

The problem requires the inclusion-exclusion principle to be applied carefully, with explicit consideration of the overlap between multiples of 10 and multiples of 15. Instead of accounting for all relevant sets and their intersections, the EASD-enhanced model mistakenly treated the exclusion step as if only the multiples of 10 needed to be subtracted from the total number of multiples of 3 or 5. In doing so, it failed to handle the overlapping elements correctly. This incomplete application of inclusion-exclusion caused some integers to be counted more than once, producing an overcount and leading to the final answer 110 rather than the correct answer 100.

Importantly, the resulting reasoning may still appear fluent and locally coherent. The word \texttt{since} introduces what seems to be a causal explanation, while \texttt{only} presents the subsequent restriction as justified and exhaustive. Together, these tokens make the flawed counting step sound more definitive than it actually is. Once this incorrect restriction is established, the later arithmetic follows naturally from it, meaning that the model does not revisit the underlying set relationships or verify whether every intersection has been handled appropriately.

This failure highlights a limitation of token-level intervention. EASD modifies tokens according to uncertainty-related signals during decoding, but a locally preferable token does not necessarily correspond to a globally correct reasoning trajectory. In highly structured tasks such as combinatorial counting, correctness depends on maintaining several logical constraints simultaneously. A token that appears reasonable in its immediate context may still encourage the model to omit a necessary case, mishandle an overlap, or prematurely simplify the problem.

Therefore, while EASD can reduce certain forms of early error propagation, it cannot guarantee error-free reasoning in every instance. The method is most effective when replacing an uncertain token redirects the model toward a more appropriate semantic or conceptual path. However, when correctness depends on a precise sequence of interdependent operations, modifying one or more tokens may introduce a different but equally plausible reasoning error. This bad case consequently illustrates both the effectiveness and the boundary of EASD: token-level correction can improve the probability of reaching a valid reasoning path, but it cannot fully replace explicit verification of global logical consistency.

\section{Hyperparameter Sensitivity Study}

We conduct a hyperparameter sensitivity study to examine how the entropy threshold and the overlap threshold affect the performance of EASD. These two thresholds jointly determine when the entropy-aware rejection mechanism is activated. The entropy threshold controls whether both the draft and target models are sufficiently uncertain at a given decoding step, while the overlap threshold measures whether the two models are uncertain in a similar region of the token space. Therefore, different threshold combinations directly influence how frequently EASD intervenes during generation.

Table~\ref{tab:hyp} reports the accuracy results across six reasoning datasets under entropy thresholds of 1, 1.5, and 2, and overlap thresholds of 0.2, 0.4, 0.6, 0.8, and 1. Overall, the results show that EASD is reasonably robust across a wide range of threshold settings. Although the absolute performance varies across datasets, most configurations remain competitive, indicating that the method is not overly dependent on a single carefully tuned hyperparameter choice.

Several trends can be observed. First, increasing the entropy threshold generally reduces the number of intervention points, because only highly uncertain decoding steps satisfy the rejection condition. This often leads to stronger average performance, suggesting that EASD benefits from focusing on more selective and high-risk tokens rather than modifying generation too aggressively. Second, the overlap threshold also plays an important role. A higher overlap threshold requires stronger agreement between the uncertain regions of the draft and target distributions, which further restricts intervention to cases where both models appear uncertain in a similar way.

The best average performance is achieved when the entropy threshold is 2 and the overlap threshold is 0.8. This setting also performs particularly well on AIME24 and GPQA, two challenging reasoning benchmarks where early token choices can strongly influence the subsequent reasoning trajectory. In contrast, AMC23 obtains its best result under a stricter overlap threshold of 1, suggesting that the optimal degree of intervention can vary slightly across datasets.

Overall, these results indicate that the default setting used in our experiments, namely $\tau_H=2$ and $\tau_O=0.8$, provides a strong trade-off between intervention frequency and output quality. It is selective enough to avoid excessive perturbation, while still allowing EASD to correct uncertain token choices that may otherwise lead to downstream reasoning errors.

\begin{table}[!h]
  \centering
  \scriptsize
  \setlength{\tabcolsep}{1.2pt}
  \renewcommand{\arraystretch}{0.92}
  \caption{EASD accuracy under different entropy thresholds $\tau_H$ and overlap thresholds $\tau_O$ (\%).}
  \label{tab:hyp}
  \begin{tabular}{ccrrrrrrr}
    \toprule
    $\tau_H$ & $\tau_O$ & Oly. & Min. & M500 & AMC & AIME & GPQA & Avg \\
    \midrule
    1   & \multirow{3}{*}{0.2} & 45.33 & 46.69 & 80.40 & 67.50 & 13.33 & 43.94 & 49.53 \\
    1.5 &                         & 46.22 & 44.85 & 82.40 & 62.50 & 13.33 & 49.49 & 49.80 \\
    2   &                         & 47.41 & 44.85 & 80.80 & 65.00 & 20.00 & 52.53 & 51.77 \\
    \midrule
    1   & \multirow{3}{*}{0.4} & 45.33 & 47.06 & 80.20 & 65.00 & 13.33 & 44.44 & 49.23 \\
    1.5 &                         & 46.22 & 44.85 & \textbf{82.40} & 62.50 & 13.33 & 50.51 & 49.97 \\
    2   &                         & 47.41 & 44.12 & 80.60 & 65.00 & 20.00 & 54.04 & 51.86 \\
    \midrule
    1   & \multirow{3}{*}{0.6} & 45.04 & \textbf{48.16} & 79.20 & 67.50 & 13.33 & 45.45 & 49.78 \\
    1.5 &                         & 46.37 & 45.59 & 82.00 & 62.50 & 13.33 & 52.02 & 50.30 \\
    2   &                         & \textbf{48.30} & 45.59 & 80.80 & 65.00 & 20.00 & 50.51 & 51.70 \\
    \midrule
    1   & \multirow{3}{*}{\textbf{0.8}} & 44.89 & 47.06 & 81.60 & 67.50 & 16.67 & 46.46 & 50.70 \\
    1.5 &                                  & 45.33 & 48.16 & 82.00 & 62.50 & 20.00 & 51.52 & 51.59 \\
    \textbf{2} &                            & 48.15 & 47.06 & 80.80 & 62.50 & \textbf{23.33} & \textbf{55.55} & \textbf{52.89} \\
    \midrule
    1   & \multirow{3}{*}{1.0} & 46.52 & 44.85 & 79.20 & \textbf{77.50} & 16.67 & 46.47 & 51.87 \\
    1.5 &                         & 47.41 & 43.75 & 81.40 & 67.50 & 16.67 & 47.48 & 50.70 \\
    2   &                         & 46.81 & 46.32 & 81.20 & 62.50 & 20.00 & 49.49 & 51.05 \\
    \bottomrule
  \end{tabular}
\end{table}

\subsection{Sensitivity to the Top-n Candidate Set Size}
\label{appendix:topn_size}

The preceding analysis varies the overlap threshold $\tau_O$ while fixing the size of the candidate sets used to compute overlap. We separately study this candidate-set size, denoted by $n$, because it determines how much of the high-probability region is used to measure draft--target agreement. A very small $n$ may omit plausible alternatives, whereas a large $n$ may include low-probability tail tokens and make the overlap signal less selective. We evaluate $n\in\{5,20,40\}$ using Qwen-2.5 (7B--32B), while keeping the remaining EASD configuration unchanged.

Table~\ref{tab:topn_size} shows that EASD outperforms SD for all three candidate-set sizes. The average score increases slightly from 52.89 at $n=5$ to 53.16 at $n=20$, before decreasing to 51.42 at $n=40$. The result at $n=20$ indicates that the default $n=5$ is not selected solely because it maximizes the reported average. More importantly, the consistent gains over SD show that the effectiveness of EASD does not depend on one carefully chosen value of $n$. We retain $n=5$ as the compact default used throughout the main experiments, since it focuses the agreement measure on the highest-probability region and avoids introducing additional low-probability candidates.

\begin{table}[!h]
  \centering
  \scriptsize
  \setlength{\tabcolsep}{1.2pt}
  \renewcommand{\arraystretch}{0.95}
  \caption{Sensitivity to the top-$n$ candidate-set size using Qwen-2.5 (7B--32B). This parameter is distinct from the overlap threshold $\tau_O$ evaluated in Table~\ref{tab:hyp}.}
  \label{tab:topn_size}
  \begin{tabular}{lrrrrrrr}
    \toprule
    Method & Oly. & Min. & M500 & AMC & AIME & GPQA & Avg. \\
    \midrule
    SD             & 45.33 & 45.22 & 80.80 & 57.50 & 13.33 & 49.49 & 48.61 \\
    EASD ($n=5$)  & \textbf{48.15} & \textbf{47.06} & 80.80 & 62.50 & \textbf{23.33} & \textbf{55.55} & 52.89 \\
    EASD ($n=20$) & 47.56 & 46.32 & \textbf{82.60} & \textbf{72.50} & 20.00 & 50.00 & \textbf{53.16} \\
    EASD ($n=40$) & 46.52 & 45.59 & 81.40 & 65.00 & 20.00 & 50.00 & 51.42 \\
    \bottomrule
  \end{tabular}
\end{table}

\section{Penalized Token Frequency Analysis}

To better understand how the entropy-aware rejection mechanism behaves under different hyperparameter settings, we further analyze the frequency of penalized tokens. Table~\ref{tab:penalty} reports the proportion of tokens affected by the penalty mechanism across the same combinations of entropy and overlap thresholds used in the hyperparameter sensitivity study. This analysis complements the accuracy results in Table~\ref{tab:hyp}, since it reveals how aggressively EASD intervenes during decoding.

The results show a clear and consistent pattern. As the entropy threshold increases, the frequency of penalized tokens decreases substantially. This is expected because a higher entropy threshold only allows intervention at decoding steps where both the draft and target models are highly uncertain. Similarly, increasing the overlap threshold also reduces the penalty frequency, because the mechanism requires stronger overlap between the top predictions of the two models. In other words, both thresholds serve as filters that control the selectivity of EASD.

This selectivity is important for maintaining stable generation quality. If the penalty mechanism is activated too frequently, it may disturb tokens that are already reasonably reliable, introducing unnecessary noise into the decoding process. Conversely, if the mechanism is activated too rarely, it may miss uncertain tokens that could trigger incorrect reasoning trajectories. The results suggest that EASD works best when it intervenes on a small but meaningful subset of tokens rather than broadly modifying the generation distribution.

The setting $\tau_H=2$ and $\tau_O=0.8$ penalizes only 1.34\% of tokens on average, yet it achieves the best average accuracy in Table~\ref{tab:hyp}. This observation supports the central design principle of EASD: effective improvement does not require frequent intervention. Instead, targeted rejection at a limited number of high-uncertainty positions can be sufficient to alter the reasoning path and prevent error propagation. The much lower penalty frequency under stricter thresholds also explains why EASD can preserve efficiency and avoid substantial disruption to the standard speculative decoding process.

Overall, the penalty frequency analysis shows that EASD provides a controllable mechanism for token-level intervention. By adjusting the entropy and overlap thresholds, one can trade off between more aggressive exploration and more conservative decoding, while the default setting offers a practical balance between correction strength, stability, and efficiency.

\begin{table}[!h]
  \centering
  \scriptsize
  \setlength{\tabcolsep}{1.2pt}
  \renewcommand{\arraystretch}{0.92}
  \caption{Frequency of penalized tokens under different entropy thresholds $\tau_H$ and overlap thresholds $\tau_O$ (\%).}
  \label{tab:penalty}
  \begin{tabular}{ccrrrrrrr}
    \toprule
    $\tau_H$ & $\tau_O$ & Oly. & Min. & M500 & AMC & AIME & GPQA & Avg \\
    \midrule
    1   & \multirow{3}{*}{0.2} & 7.69 & 8.98 & 6.97 & 8.05 & 9.85 & 26.34 & 11.31 \\
    1.5 &                         & 3.05 & 4.18 & 2.70 & 3.38 & 4.91 & 15.10 & 5.55 \\
    2   &                         & 1.11 & 1.71 & 0.91 & 1.30 & 1.57 & 8.47  & 2.51 \\
    \midrule
    1   & \multirow{3}{*}{0.4} & 7.65 & 8.85 & 6.99 & 7.98 & 9.97 & 25.87 & 11.22 \\
    1.5 &                         & 3.03 & 4.14 & 2.70 & 3.37 & 4.91 & 15.03 & 5.53 \\
    2   &                         & 1.10 & 1.67 & 0.91 & 1.29 & 1.57 & 8.73  & 2.55 \\
    \midrule
    1   & \multirow{3}{*}{0.6} & 7.42 & 8.61 & 6.66 & 7.69 & 9.32 & 22.96 & 10.44 \\
    1.5 &                         & 2.89 & 3.83 & 2.62 & 3.14 & 4.37 & 13.51 & 5.06 \\
    2   &                         & 1.01 & 1.49 & 0.87 & 1.18 & 1.43 & 6.94  & 2.15 \\
    \midrule
    1   & \multirow{3}{*}{\textbf{0.8}} & 5.79 & 6.73 & 5.67 & 6.13 & 7.53 & 14.28 & 7.69 \\
    1.5 &                                  & 2.16 & 3.00 & 2.02 & 2.34 & 3.05 & 8.11  & 3.45 \\
    \textbf{2} &                            & 0.70 & 0.97 & 0.63 & 0.77 & 0.97 & 3.97  & \textbf{1.34} \\
    \midrule
    1   & \multirow{3}{*}{1.0} & 1.98 & 2.40 & 2.13 & 1.94 & 2.61 & 3.38 & 2.41 \\
    1.5 &                         & 0.70 & 0.94 & 0.77 & 0.73 & 0.87 & 1.81 & 0.97 \\
    2   &                         & 0.18 & 0.26 & 0.19 & 0.19 & 0.22 & 0.72 & 0.29 \\
    \bottomrule
  \end{tabular}
\end{table}

\end{document}